\pdfoutput=1

\documentclass[11pt]{article}

\usepackage[preprint]{acl}

\usepackage{times}
\usepackage{amssymb} 
\usepackage{amsmath}

\usepackage{algorithm}
\usepackage{algpseudocode}
\usepackage{bbding}
\usepackage{pifont} 
\usepackage{latexsym}
\usepackage{booktabs}
\usepackage[T1]{fontenc}

\usepackage[utf8]{inputenc}
\usepackage{multirow}
\usepackage{colortbl}
\usepackage{microtype}

\usepackage{inconsolata}

\usepackage{graphicx}

\title{\includegraphics[width=0.04\textwidth]{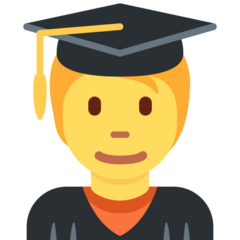}CLASS: Enhancing Cross-Modal Text-Molecule Retrieval Performance and Training Efficiency}

\author{Hongyan Wu\textsuperscript{1}, Peijian Zeng\textsuperscript{2}, Weixiong Zheng\textsuperscript{2}, Lianxi Wang\textsuperscript{3}, \\ \textbf{Nankai Lin\textsuperscript{3,\Envelope}}, \textbf{Shengyi Jiang\textsuperscript{4}}, \textbf{Aimin Yang\textsuperscript{2}}\\
    \textsuperscript{1} College of Computer, National University of Defense Technology\\
    \textsuperscript{2} School of Computer Science and Technology, Guangdong University of Technology \\
    \textsuperscript{3} School of Information Science and Technology, Guangdong University of Foreign Studies \\
    \textsuperscript{4} School of Information Technology and Engineering, Guangzhou College of Commerce\\
     \small{
   \textbf{Correspondence:} \href{mailto:email@domain}{neakail@outlook.com}
 }
        }
        
\begin{document}
\maketitle
\begin{abstract}
Cross-modal text-molecule retrieval task bridges molecule structures and natural language descriptions. Existing methods predominantly focus on aligning text modality and molecule modality, yet they overlook adaptively adjusting the learning states at different training stages and enhancing training efficiency. To tackle these challenges, this paper proposes a \textbf{C}urriculum \textbf{L}earning-b\textbf{A}sed cro\textbf{SS}-modal text-molecule training framework (\textbf{CLASS}), which can be integrated with any backbone to yield promising performance improvement. Specifically, we quantify the sample difficulty considering both text modality and molecule modality, and design a sample scheduler to introduce training samples via an easy-to-difficult paradigm as the training advances, remarkably reducing the scale of training samples at the early stage of training and improving training efficiency. Moreover, we introduce adaptive intensity learning to increase the training intensity as the training progresses, which adaptively controls the learning intensity across all curriculum stages. Experimental results on the ChEBI-20 dataset demonstrate that our proposed method gains superior performance, simultaneously achieving prominent time savings. 
\end{abstract}


\section{Introduction}

\begin{figure}[t]
  \centering
  \includegraphics[width=0.48\textwidth]{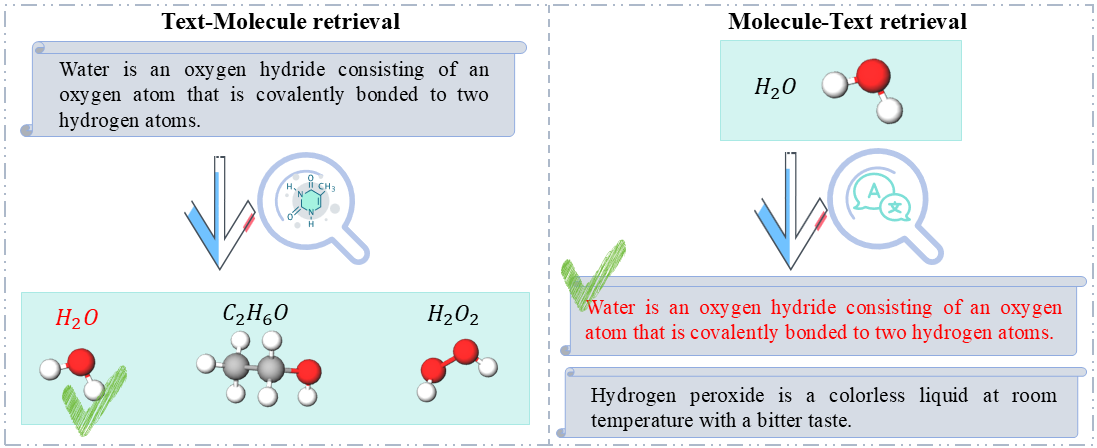}
\caption{An overview of cross-modal text-molecule retrieval tasks, involving text-molecule retrieval task and molecule-text retrieval task. The text-molecule retrieval task refers to retrieving the correct molecule using text as a query, while the molecule-text retrieval task is to retrieve the corresponding text description for the molecule. The red text indicates the ground-truth retrieval result.}
\vspace{-5pt}
\label{figure:task}
\end{figure}

In the process of drug development, pharmacologists often need to search for existing molecular structures to design new molecules or study newly discovered compounds. However, with the rapid development of bioinformatics, the scale of molecular structures currently discovered has become increasingly large, making it difficult for pharmacologists to directly and quickly retrieve the required molecular structures. As shown in Figure \ref{figure:task}, the cross-modal text-molecule retrieval task \cite{edwards-etal-2021-text2mol} consists of text-molecule retrieval task and molecule-text retrieval task, aiming to utilize text description to search for relevant molecular structures or retrieve corresponding text information using molecules as queries, enabling pharmacologists to obtain compound information more effectively. Therefore, developing a cross-modal text-molecule retrieval model is of great significance for the study of molecular structures.

Early cross-modal text-molecule retrieval work mainly focuses on optimizing text representations and molecule representations, and then calculating text-molecule similarity for retrieval. For example, some studies \cite{zeng2022deep,edwards-etal-2022-translation,liu-etal-2023-molxpt} adopt pre-trained models for encoding SMILES sequences of molecules\cite{weininger1988smiles} and text sequences. Alternatively, some studies \cite{su2022molecularmultimodalfoundationmodel,liu2023multi} represent molecules as 2D topological graphs, and then use cross-modal contrastive learning to align molecule graphs and text descriptions within a shared semantic space. 

The latest work attempts to apply methods such as adversarial learning \cite{10063974}, memory networks \cite{10821722}, and optimal transport \cite{10822800} to align text modality and molecule modality. Although these methods significantly enhance the performance of the cross-modal text-molecule retrieval task, they also increase the complexity of the models, resulting in numerous training. Furthermore, existing models lack dynamic awareness of sample difficulty and their learning states during the training process.

To address the above issues, this paper proposes a \textbf{C}urriculum \textbf{L}earning-b\textbf{A}sed cro\textbf{SS}-modal molecule training framework (\textbf{CLASS}), which combines with existing models to enhance the performance and training efficiency. Specifically, we calculate the similarity of each sample between the other samples from the training set regarding the text and molecule modalities, then counting the number of similar samples for each sample. We define the difficulty of each sample as the scale of similar samples in the training set. As is known, the training sample with the greater number of similar samples poses a greater challenge to models when retrieving. Thereafter, we gradually introduce training samples from easy to difficult at different curriculum learning stages via a sample scheduler. In addition, we propose adaptive intensity learning to coordinate learning across all curriculum stages, adaptively adjusting the learning intensity to avoid the model overfitting to simple samples. Experiments show that the method we proposed not only achieves promising performance but also effectively shortens the training time. Our contributions are summarized as follows:

1) We introduce CLASS, a cross-modal text-molecule retrieval framework compatible with any backbone, enhancing performance and training efficiency with broad applicability.

2) The CLASS framework progressively schedules training samples by difficulty, optimizing model training and reducing sample requirements.

3) We introduce adaptive intensity learning to dynamically adjust training objectives across different curriculum stages, which prevents overfitting on simple samples and improves generalization in complex scenarios.

4) Experimental results on the ChEBI-20 reveal that our method consistently surpasses other baselines, achieving prominent time savings while improving model performance.

\section{Related Work}
With the advent of multimodal representation learning, the cross-modal text-molecule retrieval task has garnered significant attention. Existing methods can be broadly categorized into three groups: statistic-based, hashing-based, and deep learning-based approaches.

\subsection{Statistic-Based Methods}
Statistic-based methods aim to bridge the modality gap by analyzing feature correlations in subspace projections. Canonical correlation analysis \cite{hardoon2004canonical}, kernel canonical correlation analysis \cite{lai2000kernel}, and deep canonical correlation analysis \cite{andrew2013deep} are representative methods that maximize correlations between text and molecule embeddings in a shared latent space. While these methods demonstrated initial success, they often struggle with complex, high-dimensional data and fail to capture deep semantic relationships between different modalities.

\subsection{Hashing-Based Methods}
Hashing-based methods focus on mapping high-dimensional multimodal features into binary codes to enable efficient similarity search. Techniques such as spectral hashing \cite{weiss2008spectral}, self-taught hashing \cite{zhang2010self}, and iterative quantization \cite{gong2012iterative} project text and molecule representations into a shared Hamming space, where retrieval is accelerated by comparing binary codes. Notably, the discretization process may overlook nuanced semantic dependencies between textual concepts and molecular structural motifs, thereby limiting retrieval accuracy.

\begin{figure*}[!htbp]
\centering
\includegraphics[scale=0.48]{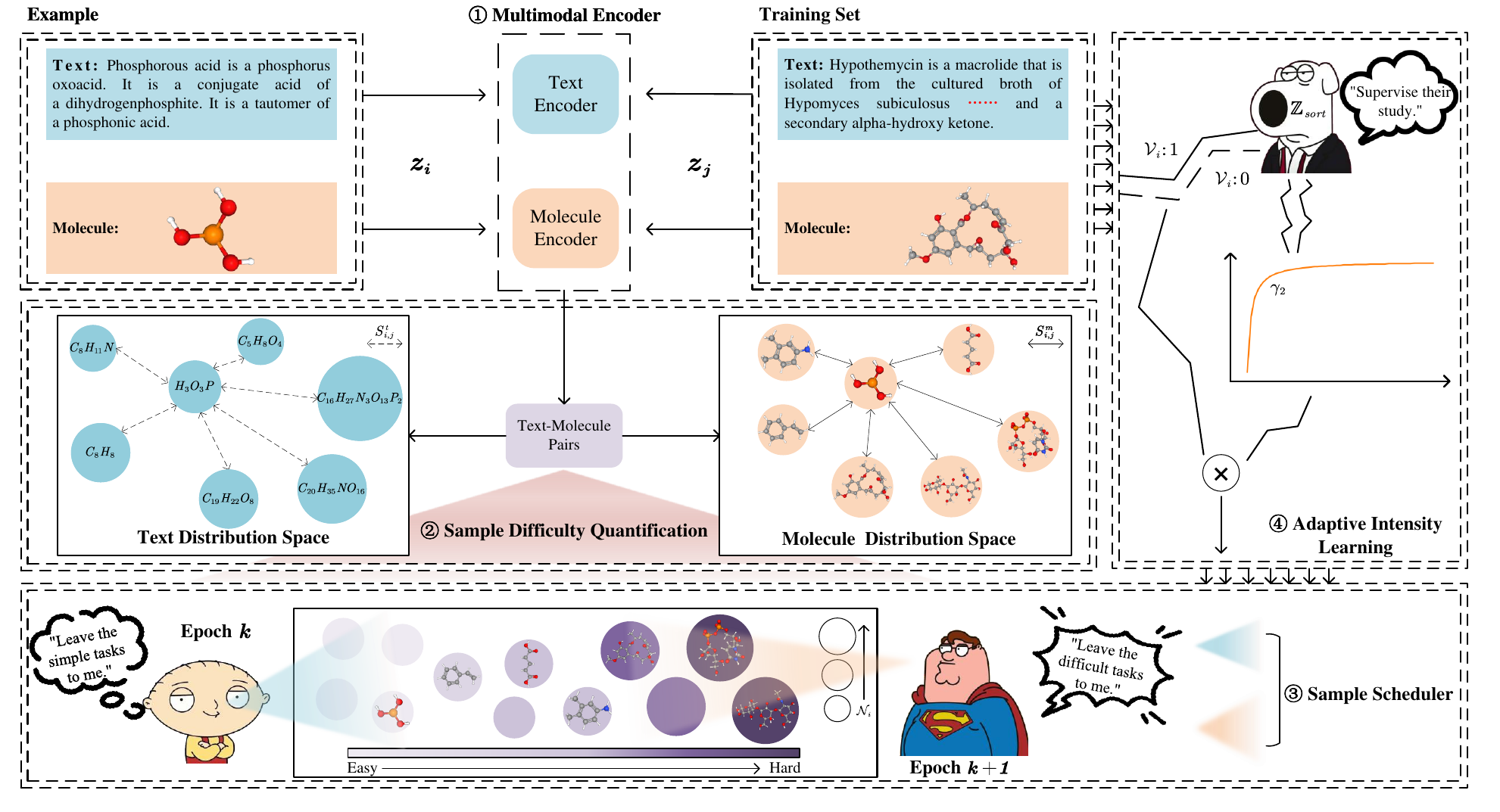}
\caption{Overview of CLASS. All the molecules are represented by white spheres for H, red for O, gray for C, and yellow for P. Initially, the \textbf{multimodal encoder} (\textcircled{1}) encodes $z_i$ and $z_j$, and then inputs them into the \textbf{sample difficulty quantification} (\textcircled{2}) to calculate the similarity between samples, quantifying the difficulty of sample $z_i$ based on the number $\mathcal{N}_{i}$ of confusing samples. Thereafter, the \textbf{sample scheduler} (\textcircled{3}) based on a curriculum learning strategy introduces training samples via an easy-to-hard paradigm. Finally, the \textbf{adaptive intensity learning} (\textcircled{4}) dynamically adjusts the model's training intensity to control the global training process of the model.}
\vspace{-5pt}
\label{fig:overview}
\end{figure*}

\subsection{Deep Learning-Based Methods}
Deep learning pays attention to mapping the semantic representations of different modalities into a shared embedding space to facilitate similarity evaluation. Adversarial networks are widely employed in image-recipe retrieval tasks \cite{wang2019learning, li2021multi}. For instance, \citet{10063974} introduces adversarial training to align text and molecule representations, achieving robust modality alignment through triplet loss and a min-max game strategy. Another line of work involves the introduction of contrastive learning, with \citet{su2022molecularmultimodalfoundationmodel} deploying contrastive learning on molecular graphs and text embeddings. Additionally, \citet{10822800} leverages optimal transport for multi-grained alignment. \citet{10821722} decomposes molecules into hierarchical graphs and aligns them with text at multiple granularities using optimal transport, significantly improving retrieval precision. However, these methods overlook the aspect of enhancing training efficiency and varying difficulty across samples.

\section{Proposed CLASS Framework}
In this section, we provide a detailed description of our CLASS framework, as illustrated in Figure \ref{fig:overview}. Initially, we utilize the \textbf{multimodal encoder} to generate text representation and molecule representation via a text encoder and a molecule encoder. Next, we use the \textbf{sample difficulty quantification} to measure the difficulty of each training sample considering the number of similar samples in the training set. We further design the \textbf{sample scheduler} to introduce samples following easy-to-difficult paradigm as the training advances. Simultaneously, we propose \textbf{adaptive intensity learning} to coordinate training across all curriculum stages, which adaptively control the learning intensity. The detailed pseudo-code of the entire algorithm is shown in the Appendix \ref{algorithm:cross_modal_retrieval}.

It is noted that our framework exhibits potent generality and flexibility. It can seamlessly adapt to any cross-modal text-molecule retrieval model, which implies that our framework is not restricted by specific models, providing a universal solution for cross-modal text-molecule retrieval tasks.

\subsection{Problem Definition}
We provide a detailed definition of cross-modal text-molecule retrieval, which involves the text-molecule retrieval task and molecule-text retrieval task. Given a text description set $\mathbb{T}$ without molecule information and a molecule set $\mathbb{M}$ without text information, it aims to retrieve the corresponding molecule from set $\mathbb{M}$ for a text instance or match the corresponding text description from set $\mathbb{T}$ for each molecule. We define $\mathbb{Z}$ as the training set, which consists of a series of text-molecule pairs, that is $\mathbb{Z} = \{z_1, z_2, \cdots, z_L\}$. Here, $z_i = (t_{i}, m_{i})$, indicating the $i$-th text-molecule pair, composed of the text description $t_{i}$ and the molecule $m_{i}$, and $L$ is the number of training samples.

\subsection{Multimodal Encoder}


We exploit SciBERT \cite{scibert} outperforming in encoding chemical texts to encode text description. More precisely, for a text description $t$, the [CLS] token is concatenate with the header of a text sequence. Then we input the sequence into the SciBERT to extract the feature vector corresponding to symbol [CLS] as the text representation $h^t$. Following the study \cite{10063974}, we employ molecule representation $h^m$ encoded by Mol2vec algorithm \cite{mol2vec}, which converts molecule graph structures into ``sentences'' of substructures.

\subsection{Sample Difficulty Quantification}
For the given text-molecule pair $z_i$, we calculate its text similarity $S_{i,j}^t$ and molecule similarity $S_{i,j}^m$ with each instance $z_j$ from the training set via cosine similarity according to Eq. \ref{text} and Eq. \ref{molecule}, obtaining the mean value $S_{i,j}$ of the similarity of the two modalities, as in Eq. \ref{mean}:
\begin{equation}
\label{text}
    S_{i,j}^t = cos(h_i^t, h_j^t),
\end{equation}
\begin{equation}
\label{molecule}
    S_{i,j}^m = cos(h_i^m, h_j^m),
\end{equation}
\begin{equation}
\label{mean}
S_{i,j} = \frac{1}{2}(S_{i,j}^t+S_{i,j}^m).
\end{equation}

Thereafter, as indicated in Eq. \ref{num} and Eq. \ref{I}, we count the number $\mathcal{N}_{i}$ of samples in the training set that are similar to sample $z_i$ with a similarity of more than a threshold value $\sigma$, indicating the difficulty level of the sample:
\begin{equation}
\label{num}
    \mathcal{N}_{i}=\sum_{j=1, j \neq i}^{|\mathbb{Z}|} \mathcal{I}_{i,j},
\end{equation}
\begin{equation}
\label{I}
    \mathcal{I}_{i,j}=\left\{\begin{array}{ll}
1, & S_{i,j}>\sigma, \\
0, &otherwise.
\end{array}\right. 
\end{equation}

It is acknowledged that samples with a larger number of similar samples pose a greater challenge to the model during training. The model will have more difficulty in learning to distinguish between these similar samples. Therefore, we sort the samples in the training set in order of the increasing number of similar samples to get the final training set $\mathbb{Z}_{sort} =\left\{z_{i}| \mathcal{N}_{i} < \mathcal{N}_{j}, z_{i} \in \mathbb{Z}, z_{j} \in \mathbb{Z}, i < j \right\}$. Intuitively, the samples in the set $\mathbb{Z}_{sort}$ simultaneously follow the order from simple to difficult. During model training, we schedule samples for training in the order of the set $\mathbb{Z}_{sort}$, which elicits the model to gradually learn and adapt to samples of different difficulty levels.



\subsection{Sample Scheduler}
Different from previous work \cite{edwards-etal-2021-text2mol}, treating all samples equally important, we design sample scheduler following easy-to-difficult paradigm based on curriculum learning strategy. Specifically, curriculum learning is integrated into the total training epoch to facilitate model retrieval learning. During the training process in curriculum learning, samples in the training set $\mathbb{Z}_{sort}$ are gradually scheduled from easy to difficult as the pace parameter $k$-th increases. In the $k$-th epoch, the proportion of samples used for training is
\begin{equation}
\label{ratio}
    \lambda=\alpha+\beta \cdot k,
\end{equation}
where $\lambda$ denotes the proportion of samples introduced in the $k$-th epoch, $\alpha$ and $\beta$ are the initial ratio of training samples and growth factor. We conduct experimental investigations in different $\alpha$ and $\beta$, as presented in Appendix \ref{app:Parameter Exploration}.  As a result, the training samples used for model learning at each epoch stage consist of the set $\mathbb{D}_k=\left\{z_i | z_i \in \mathbb{Z}_{sort}, i \leq \lambda |\mathbb{Z}_{sort}|\right\}$. After all training samples are introduced for model learning, namely $|\mathbb{Z}_{sort}|=|\mathbb{D}_k|$, model continues to be optimized based on all the samples at remaining phase.


\subsection{Adaptive Intensity Learning}
During the model training process, we implement adaptive intensity learning via curriculum learning strategy. At each curriculum stage $k$, we control the complexity of the training sample, and concurrently modify the training objective to dynamically adjust the degree of model training. In the initial curriculum stage, when the model learns simple samples, the training intensity of the model is appropriately reduced to avoid overfitting. As the curriculum advances to the complex sample learning stage, the model training intensity is gradually increased to induce the model to better learn complex modal features, thus improving the model's performance and generalization ability in complex scenarios. Notably, our framework can be combined with any backbone that adopts methods such as contrastive learning or adversarial learning \cite{10063974, 10821722, 10822800}, yielding promising improvement for model's performance. 

More precisely, we design the two learning intensity curves in Eq. \ref{gamma1} and Eq. \ref{gamma2} to achieve adaptive curriculum learning for the training of the model with the increase of the epoch $k$:
\begin{equation}
\label{gamma1}
    \gamma_{1}=\frac{1}{1+e^{-k-1}},
\end{equation}
\begin{equation}
\label{gamma2}
    \gamma_{2}=\frac{k}{1+k}.
\end{equation}

We investigate the effect of the two learning intensity curves on model performance in Subsection \ref{learning intensity curves}. Then, the training objective of any model is adaptively adjusted according to the learning intensity curve $\gamma\in \{\gamma_{1}, \gamma_{2}\}$. As shown in Eq. \ref{l_epoch}, within a curriculum learning stage $k$, we define the training objective of the model as $\mathcal{L}_{epoch}$, computed by the loss $\mathcal{L}$ of each introduced training sample via sample scheduler. For one sample $z_i$, $\mathcal{V}_i=1$ means that $z_i$ is selected to train the model in the epoch  $k$, whereas $\mathcal{V}_i=0$ represents that $z_i$ is unselected in the epoch $k$. The strategy ensures that the model is trained at low intensity during the curriculum learning of simple samples phase, and gradually increases the training intensity as the complexity of the training samples increases:

\begin{equation}
\label{l_epoch}
    \mathcal{L}_{epoch}=\gamma (\sum_{i=1}^{|\mathbb{Z}_{sort}|} \mathcal{V}_{i} \cdot \mathcal{L}),
\end{equation}

\begin{equation}
\label{V}
    \mathcal{V}_i=\left\{\begin{array}{ll}
1, & i \leq \lambda |\mathbb{Z}_{sort}|, \\
0, & otherwise.
\end{array}\right. 
\end{equation}

\begin{table*}[!ht]
    \centering
    \small
    \caption{Main results of our method and baselines on the ChEBI-20 dataset for the text-molecule retrieval task. The best results are shown in bold. $\uparrow$ denotes that the higher is the better. $^{\dagger}$ represents our reproduced results of AMAN and ORMA and the other results are reported in the previous works.}
    \begin{tabular}{c|cccc}
    \toprule
        \textbf{Models} & \textbf{Hits@1($\uparrow$)} & \textbf{Hits@10($\uparrow$)}
        & \textbf{MRR($\uparrow$)} & \textbf{Mean Rank($\downarrow$)} \\ 
        \midrule
        MLP-Ensemble \cite{edwards-etal-2021-text2mol} & 29.4\% & 77.6\% & 0.452 & 20.78 \\
        GCN-Ensemble \cite{edwards-etal-2021-text2mol} & 29.4\% & 77.1\% & 0.447 & 28.77 \\
        All-Ensemble \cite{edwards-etal-2021-text2mol} & 34.4\% & 81.1\% & 0.499 & 20.21 \\
        MLP+Atten \cite{edwards-etal-2021-text2mol} & 22.8\% & 68.7\% & 0.375 & 30.37 \\
        MLP+FPG \cite{edwards-etal-2021-text2mol} & 22.6\% & 68.6\% & 0.374 & 30.37  \\
        Atomas-base \cite{atmos} & 50.1\% & 92.1\% & 0.653 & 14.49 \\
        Memory Bank \cite{10821722} & 56.5\% & 94.1\% & 0.702 & 12.66 \\
        \midrule
        AMAN$^{\dagger}$ \cite{10063974} & 49.2\% & 91.6\% & 0.646 & 20.15 \\
        \rowcolor{cyan!10} \textbf{CLASS (AMAN)} & \textbf{51.1\%} & \textbf{92.6}\% & \textbf{0.660} & \textbf{16.80} \\
        \midrule
        ORMA$^{\dagger}$ \cite{10822800} & 66.4\% & \textbf{93.8\%} & \textbf{0.776} & 19.58 \\
        \rowcolor{cyan!10} \textbf{CLASS (ORMA)} & \textbf{67.4\%} & 93.4\% & 0.774 & \textbf{17.82} \\
    \bottomrule
    \end{tabular}
    \vspace{-5pt}
    \label{tab:text2mol}
\end{table*}

\section{Experiments}
\subsection{Experiment Setting}
\paragraph{Dataset.} We conduct experiments on the ChEBI-20 dataset \cite{edwards-etal-2021-text2mol}, which is collected from PubChem \cite{pubchem} and Chemical Entities of Biological Interest (ChEBI) \cite{bio}. The ChEBI-20 dataset consists of 26,408 training samples, 3301 validation samples and 3301 test samples, with a total of 33,010 pairs. 

\paragraph{Evaluation Metrics.} Our evaluation involves text-molecule retrieval task and molecule-text retrieval task. Consistent with prior studies \cite{edwards-etal-2021-text2mol, 10063974}, owing to one-to-one corresponding for the text and molecule in this dataset, we employ Mean Reciprocal Rank (MRR),  Mean Rank, Hits@1, and Hits@10 as evaluation metrics. More detailed descriptions of metrics are provided in Appendix \ref{app:Evaluation Metrics}.

\paragraph{Comparison Methods.} We compare with the following methods to demonstrate the superiority of our methods, including \textbf{MLP-Ensemble} \cite{edwards-etal-2021-text2mol}, \textbf{GCN-Ensemble} \cite{edwards-etal-2021-text2mol}, \textbf{All-Ensemble} \cite{edwards-etal-2021-text2mol}, \textbf{MLP+Atten} \cite{edwards-etal-2021-text2mol}, \textbf{MLP+FPG} \cite{mlp_fpg}, \textbf{Atomas-base} \cite{atmos}, \textbf{Memory Bank} \cite{10821722}, \textbf{AMAN} \cite{10063974} and \textbf{ORMA} \cite{10822800}. More detailed descriptions of baselines are provided in Appendix \ref{app:Comparison Methods}.

\paragraph{Implementation.} We conduct all experiments using the A100 GPU. The SciBERT \cite{scibert} is used as a text encoder to learn the representation with the dimension of 300, and the Mol2Vec \cite{mol2vec} is used as a molecule encoder to produce the molecule representations. We choose the AMAN model and ORMA model as our CLASS framework's backbones. We consistently adopt the Adam optimizer \cite{adam} with a fixed initial learning rate of 1e-4 across all experiments. Training consists of 60 epochs with a batch size of 32 for both training and evaluation. Additionally, we conduct a grid search to optimize the model's hyperparameters. While using the AMAN model as the backbone, the trade-off hyperparameters of $\alpha$ and $\beta$ are set to 40 and 3 in text-molecule retrieval task, then for molecule-text retrieval task with these hyperparameters set to 20 and 4. When the ORMA model is adopted as the backbone, the $\alpha$ and $\beta$ are set to 70 and 3 in text-molecule retrieval task, and in the molecule-text retrieval task, they are set to 60 and 2. The similarity threshold used to measure the difficulty of the sample $\delta$ is set to 0.99.

\begin{table*}[!ht]
    \centering
    \small
    \caption{Main results of our method and baselines on the ChEBI-20 dataset for the molecule-text retrieval task. The best results are shown in bold. $\uparrow$ denotes that the higher is the better. $^{\dagger}$ represents our reproduced results of AMAN and ORMA and the other results are reported in the previous works.}
    \begin{tabular}{c|cccc}
    \toprule
        \textbf{Models} & \textbf{Hits@1($\uparrow$)} & \textbf{Hits@10($\uparrow$)}
        & \textbf{MRR($\uparrow$)} & \textbf{Mean Rank($\downarrow$)} \\ 
        \midrule
        All-Ensemble \cite{edwards-etal-2021-text2mol} & 25.2\% & 74.1\% & 0.408 & 21.77 \\
        Atomas-base \cite{atmos} & 45.6\% & 90.3\% & 0.614 & 15.12 \\
        Memory Bank \cite{10821722} & 52.3\% & 93.3\% & 0.673 & 12.29 \\
	\midrule
        AMAN$^{\dagger}$ \cite{10063974} & 46.2\% & 91.1\% & 0.622 & 19.93 \\
        \rowcolor{cyan!10} \textbf{CLASS (AMAN)} & \textbf{47.8\%} & \textbf{91.6\%} & \textbf{0.631} & \textbf{15.14} \\ 
        \midrule
        ORMA$^{\dagger}$ \cite{10822800} & 60.9\% & \textbf{93.2\%} & 0.733 & \textbf{10.71} \\
        \rowcolor{cyan!10} \textbf{CLASS (ORMA)} & \textbf{62.0\%} & 92.7\% & \textbf{0.738} & 14.59 \\
    \bottomrule
    \end{tabular}
    \vspace{-5pt}
    \label{tab:mol2text}
\end{table*}

\begin{table*}[!ht]
    \centering
    \small
    \caption{Ablation study of AMAN and our method regarding three dimensions: ablating text similarity or removing molecule similarity for sample difficulty quantification, and ablating dynamic loss. \ding{51} indicates using the strategy when training the model. 
    }
    \begin{tabular}{ccc|cccc}
    \toprule
        \multicolumn{7}{c}{\textbf{Text-Molecule Retrieval}} \\
        \textbf{Text} & \textbf{Molecule} & \textbf{\begin{tabular}[c]{@{}c@{}}Adaptive \\ Intensity Learning\end{tabular}} & \textbf{Hits@1($\uparrow$)} & \textbf{Hits@10($\uparrow$)} & \textbf{MRR($\uparrow$)} & \textbf{Mean Rank($\downarrow$)} \\ 
        \midrule
        \ding{51} & \ding{51} & \ding{51} & \textbf{51.1\%} & \textbf{92.6\%} & \textbf{0.660} & 16.80 \\
        & \ding{51} & \ding{51} & 50.5\% & 92.5\% & 0.654 & 20.32 \\
        \ding{51} & & \ding{51} & 49.2\% & 91.9\% & 0.643 & 17.06 \\
        \ding{51} & \ding{51} & & 50.3\% & 92.5\% & 0.651 & \textbf{13.44} \\
        \midrule
        \multicolumn{7}{c}{\textbf{Molecule-Text Retrieval}} \\
        \textbf{Text} & \textbf{Molecule} & \textbf{\begin{tabular}[c]{@{}c@{}}Adaptive \\ Intensity Learning\end{tabular}} & \textbf{Hits@1($\uparrow$)} & \textbf{Hits@10($\uparrow$)} & \textbf{MRR($\uparrow$)} & \textbf{Mean Rank($\downarrow$)} \\ 
        \midrule
        \ding{51} & \ding{51} & \ding{51} & \textbf{47.8\%} & 91.6\% & \textbf{0.631} & 15.14 \\
        & \ding{51} & \ding{51} & 47.0\% & 91.3\% & 0.626 & \textbf{14.88} \\
        \ding{51} & & \ding{51} & 46.2\% & 90.8\% & 0.617 & 16.26 \\
        \ding{51} & \ding{51} & & 45.5\% & \textbf{92.2\%} & 0.618 & 16.30 \\
    \bottomrule
    \end{tabular}
    \vspace{-5pt}
    \label{tab:abla}
\end{table*}

\subsection{Overall Results}
As illustrated in Table \ref{tab:text2mol} and Table \ref{tab:mol2text}, models employing our methods yield remarkable improvement compared with baselines across all evaluation metrics for both two retrieval tasks. 

In the task of text-molecule retrieval, our methods CLASS (AMAN) and CLASS (ORMA) consistently outperform baselines of AMAN and ORMA, yielding an improvement of 1.0\%-1.9\% in Hits@1 and a reduction of 1.76-3.35 concerning Mean Rank, which demonstrates that our methods can be combined with any backbone to enhance the performance of retrieval learning. Notably, AMAN integrating our methods yields improved performance in Hits@10 and MRR, concurrently achieving considerable gains of 1.0\%, 0.5\% and 0.007 over the potent baseline Atomas-base in Hits@1, Hits@10 and MRR. These results indicate that our model can not only retrieve the matching answer for more instances but also improve the sort of retrieval results. Although the results of CLASS (ORMA) are slightly lower in Hits@10 and MRR, it is competitively close to the best-performing ORMA.   


\begin{table*}[!ht]
    \centering
    \small
    \caption{The effect of different learning intensity curves in adaptive intensity learning based on CLASS (AMAN) for text-molecule rerieval task. The best results are shown in bold.}
    \begin{tabular}{c|cccc}
    \toprule
        \textbf{Learning Intensity Curve} & \textbf{Hits@1($\uparrow$)} & \textbf{Hits@10($\uparrow$)}
        & \textbf{MRR($\uparrow$)} & \textbf{Mean Rank($\downarrow$)} \\ 
        \midrule
        $\gamma_{1}$ & 50.5\% & 91.6\% & 0.653 & 20.50  \\
        $\gamma_{2}$ & \textbf{51.1}\% & \textbf{92.6}\% & \textbf{0.660} & \textbf{16.80}  \\
    \bottomrule
    \end{tabular}
    \vspace{-5pt}
    \label{tab:loss weight}
\end{table*}

In the task of molecule-text retrieval, AMAN and ORMA equipped with our methods demonstrate superior performance, leading in Hits@1 and MRR with an improvement of 1.1\%-1.6\% and 0.005-0.009, respectively. The consistent enhanced performance proves the universality of CLASS. Meanwhile, the outstanding performance of CLASS (ORMA) across all metrics highlights the potential of our model to surpass generalized baselines in the molecule-text retrieval task.


\subsection{Ablation Study}
As presented in Table \ref{tab:abla}, to investigate the contribution of sample scheduler and adaptive intensity learning, we ablate different strategies based on CLASS (AMAN) to conduct experiments in both two retrieval tasks. When discarding the text similarity for quantifying the difficulty of sample in the sample scheduler, the performance of the model shows varying degrees of degradation in both two retrieval tasks. Meanwhile, it is obvious that ablating molecule similarity for quantifying the difficulty of sample results in drops by 1.6\%-1.9\%, 0.7\%-0.8\%, 0.02-0.014 and 0.26-1.12 across all metrics in model's performance correspondingly. The results indicate that combining the similarity of the two modalities when quantifying the difficulty of the samples can be effective, contributing to a higher utility in curriculum learning. Eliminating the adaptive intensity learning strategy yields considerable performance declines in Hits@1 and MRR, with decreases of 0.8\%-2.3\% and 0.013-0.009, showing the effectiveness of controlling training intensity. Adaptive intensity learning is conducive to learning multimodal representations by training the model adaptively at different stages.

\subsection{Analysis}
\begin{figure}[t]
  \centering
  \includegraphics[width=0.47\textwidth]{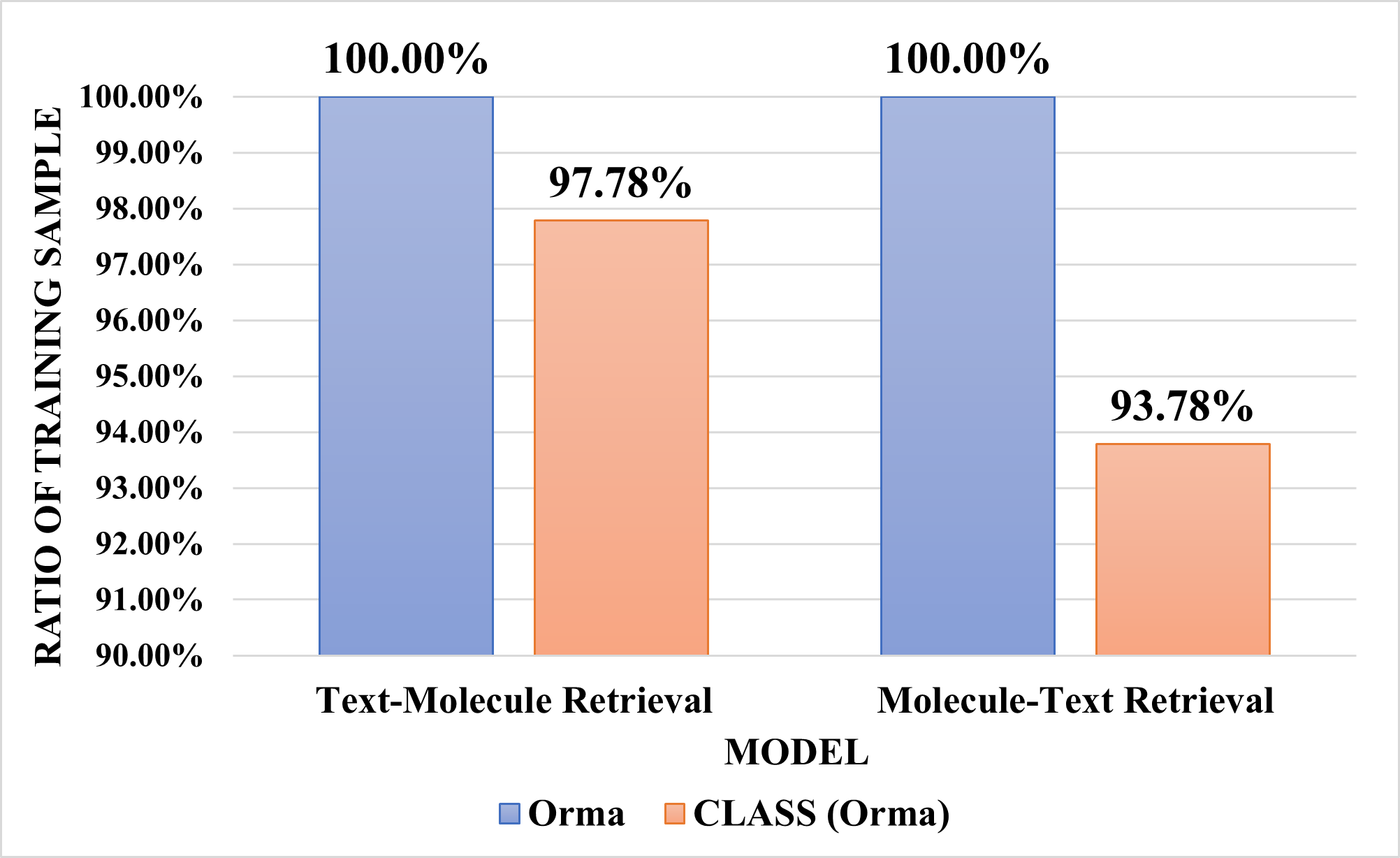}
\caption{The ratio of training sample used in ORMA and our CLASS (ORMA) during total training epochs for both two retrieval tasks.}
\vspace{-5pt}
\label{figure:orma}
\end{figure}

\begin{figure}[t]
  \centering
  \includegraphics[width=0.47\textwidth]{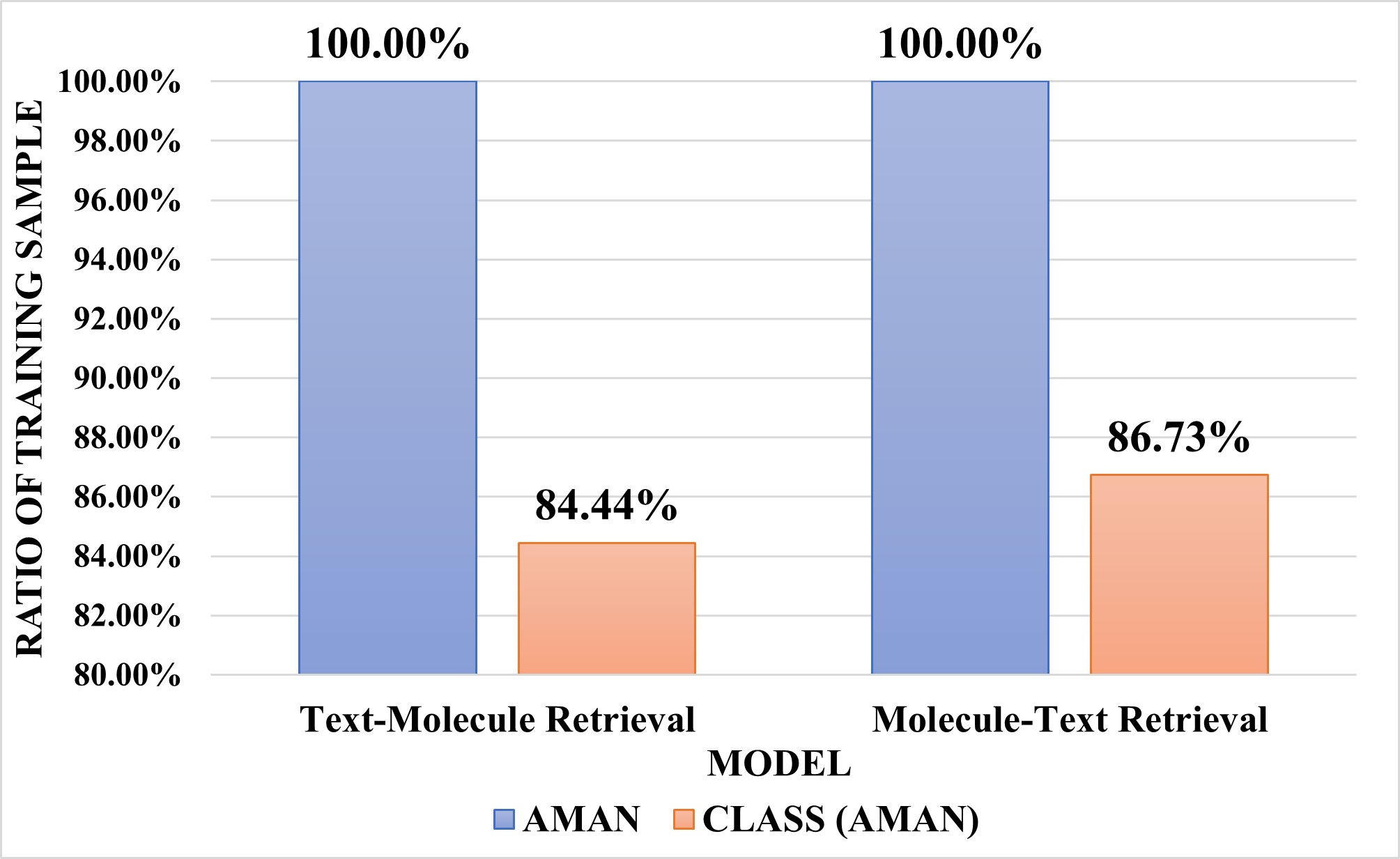}
\caption{The ratio of training sample used in AMAN and our CLASS (AMAN) during total training epochs for both two retrieval tasks.}
\vspace{-5pt}
\label{figure:aman}
\end{figure}

\paragraph{Analysis of Training Efficiency.} As illustrated in Figure \ref{figure:orma} and \ref{figure:aman}, we calculate the ratio of training data used in ORMA and AMAN during total training epochs, to demonstrate that our method yields notable time and cost savings. The results show that our method consistently reduces the scale of training samples compared to baselines. Notably, the CLASS method achieves prominent improvements in improving the training efficiency of the AMAN model, solely using a sample size of 84.44\% in the text-molecule retrieval task and a sample size of 86.73\% in the molecule-text retrieval task. The time cost savings benefit from the sample scheduler, which does not require learning all the training samples in the initial phase, but rather gradually introduces them in a step-by-step manner.

\begin{figure*}[t]
  \centering
  \includegraphics[width=1\textwidth]{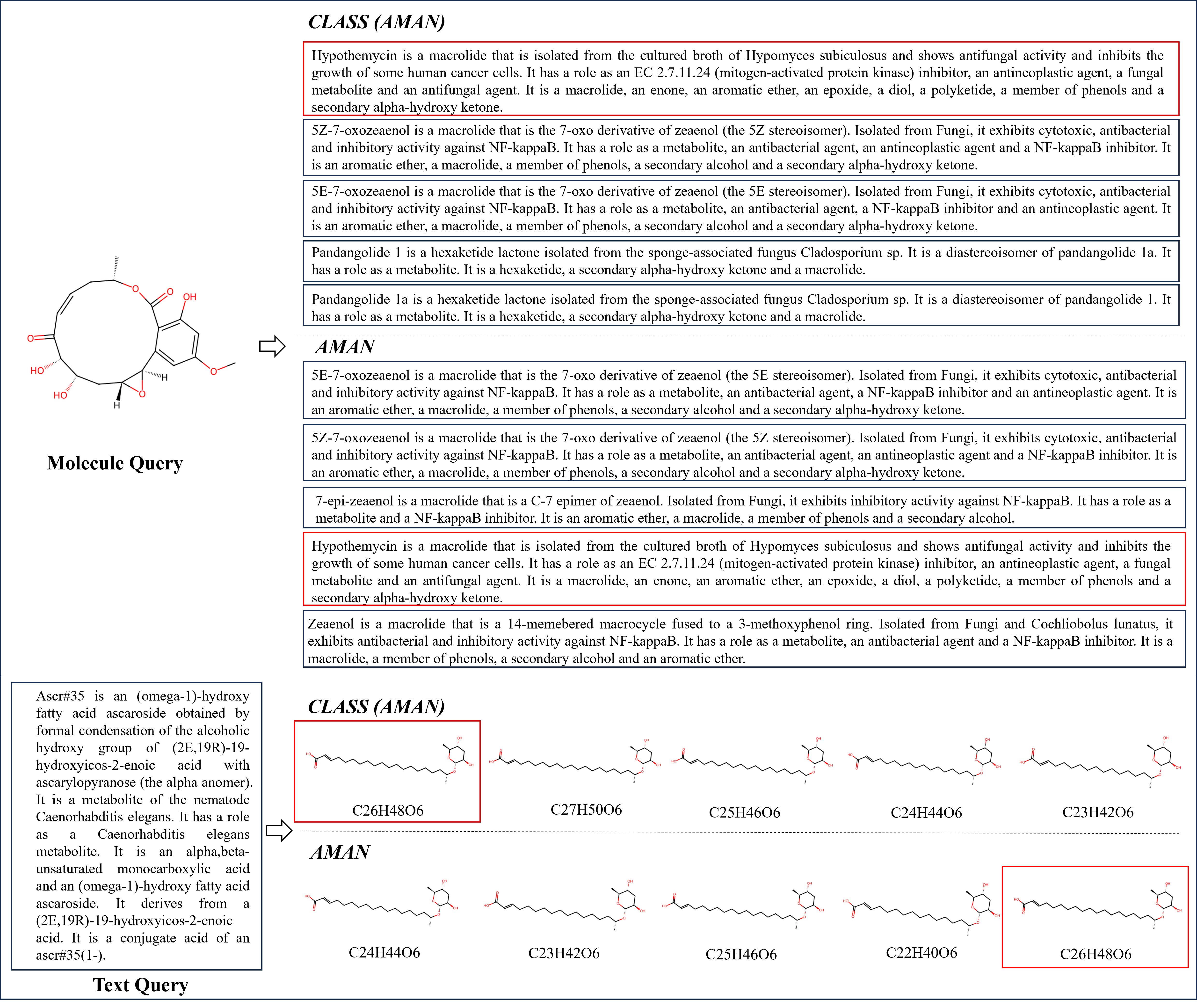}
\caption{Cases of our model and AMAN in the cross-modal text-molecule retrieval task. The red box highlights the candidate retrieved by our model that exactly corresponds to the ground truth.}
\vspace{-5pt}
\label{figure:case}
\end{figure*}

\paragraph{Impact of Learning Intensity Curves.}
\label{learning intensity curves}
As shown in Eq. \ref{gamma1} and Eq. \ref{gamma2}, we harness the sigmoid function and the inverse proportion function as the learning intensity curve, respectively, and explore the impact of the two strategies based on the AMAN model. Specifically, when adjusting the training objective of the model based on Eq. \ref{gamma1} and Eq. \ref{gamma2}, the value of both $\gamma_{1}$ and $\gamma_{2}$ gradually increase with the increasing epoch, implying that the training of the model is strengthened in the complex sample learning stage. This is consistent with the idea of curriculum learning, where the model training intensity is appropriately reduced in the curriculum learning stage for simple samples to avoid overfitting with simple samples, while the complex stage adaptively increases the loss weights to enhance the learning of features for difficult samples. As shown in Table \ref{tab:loss weight}, the model achieves better performance with $\gamma_{2}$ as the learning intensity curve, demonstrating the potential of guiding the training of the model via curriculum learning. This may be attributed to the fact that $\gamma_{1}$ increases faster and reaches the convergence process earlier, whereas $\gamma_{2}$  changes more uniformly throughout the process and reaches the state of convergence later, which contributes to coordinating the model training across all curriculum phase.

\subsection{Case Study}
As illustrated in Figure \ref{figure:case}, we provide some typical cases from test set in terms of CLASS (AMAN) and important baseline AMAN to further validate the effectiveness of our methods in cross-modal text-molecule retrieval task. Retrieved with a text query and a molecule query, respectively, our curriculum-learning-based methods yield ground truth as the top-1 candidate in both tasks, while AMAN fails. It is consistent with experimental results that our approach significantly improves the Hits@1 metric. More importantly, in text-molecule retrieval, although the visualizations of candidate molecules show remarkably similar, our method accurately retrieves ground molecules as the top-1 retrieval result. This demonstrates the superiority of our method in learning the retrieval of complex samples, proving the strong potential of aligning cross-modal representations through adaptive training.

\section{Conclusion}
In this paper, we propose a cross-modal text-molecule training framework CLASS, involving a sample scheduler and adaptive intensity learning. We quantify sample difficulty and schedule training samples from easy to difficult. Additionally, adaptive intensity learning coordinates training across stages, dynamically adjusting training intensity. Extensive experiments reveal that our method significantly improves the performance of the model in cross-modal text-molecule retrieval tasks, simultaneously achieving prominent time savings. More importantly, our framework can be integrated with any backbone to gain promising improvement in performance, with strong versatility. 

In the future, we will explore a more effective learning intensity curve for adaptive intensity learning. Subsequently, we extend the adaptive intensity learning to adaptively focus on adjusting different learning tasks in training objectives during various stages. Moreover, despite the effectiveness of the molecular similarity calculation method we used, we plan to investigate more potent molecular similarity calculation methods.

\section*{Limitations}
Despite the promising results of the proposed CLASS framework in cross-modal text-molecule retrieval, several limitations exist. While the adaptive intensity learning curves used in this study have demonstrated certain effects, there may be more sophisticated functions that can better adapt to the model's learning process at different stages. In terms of the molecular similarity calculation, the method used in this work may not be able to fully capture the intricate structural and chemical properties of molecules, which could impact the quality of cross-modal alignment and retrieval performance.

\section*{Ethics Statement}
The datasets and pre-trained language models used in our study come from open-access repositories. This ensures that we comply with all relevant ethical standards and authorizations. We strictly follow established research ethics throughout our research.

\bibliography{custom}

\clearpage

\appendix

\section{Pseudocode of Our Proposed Method}

The pseudocode of the implementation process of our proposed method CLASS is shown in Algorithm \ref{algorithm:cross_modal_retrieval}.

\begin{algorithm}[!htb]
\caption{Curriculum Learning-based Cross-modal Text-Molecule Training Framework}
\label{algorithm:cross_modal_retrieval}
\begin{algorithmic}[1] 
    \State \textbf{Input}:  \\
    \quad Similarity threshold $\sigma$ \\
    \quad Initial number of samples $\alpha$ \\
    \quad Growth factor $\beta$ \\
    \quad Learning intensity curve $\gamma$ \\
    \quad Total epochs $E$
    \State \textbf{Output}: Trained model
    \State Encode texts using SciBERT to generate token representations $h^t$
    \State Encode molecules using Mol2Vec to generate molecule representations $h^m$
    \For {each text-molecule pair $z_i$}
        \State Calculate text ($h^t$) similarity $S_{i,j}^t$ and molecule ($h^m$) similarity $S_{i,j}^m$ with cosine similarity
        \State Compute mean similarity $S_{i,j} = \frac{S_{i,j}^t + S_{i,j}^m}{2}$
        \State Count similar samples $\mathcal{N}_{i}=\sum_{j=1, j \neq i}^{|\mathbb{Z}|} \mathcal{I}_{i,j}$, where $\mathcal{I}_{i,j} = 1$ if $S_{i,j} > \sigma$ else $0$
    \EndFor
    \State Sort the training set $\mathbb{Z}$ by increasing $\mathcal{N}_{i}$ as $\mathbb{Z}_{sort}$ (from simple to difficult)
    \For {epoch $k = 1$ to $E$}
        \State Determine $\lambda = \alpha + \beta \cdot k$
        \State Select $\lambda |\mathbb{Z}_{sort}|$ samples from $\mathbb{Z}_{sort}$ as $\mathbb{D}_k$
        \For {each sample $z_i$ in $\mathbb{D}_k$}
            \If {$i \leq \lambda |\mathbb{Z}_{sort}|$}
                \State Set $\mathcal{V}_i = 1$
            \Else
                \State Set $\mathcal{V}_i = 0$
            \EndIf
            \State Calculate epoch loss $\mathcal{L}_{epoch} = \gamma (\sum_{i=1}^{|\mathbb{Z}_{sort}|} \mathcal{V}_{i} \cdot \mathcal{L})$
        \EndFor
    \EndFor
    \State \textbf{return} Trained model
\end{algorithmic}
\end{algorithm}

\section{Evaluation Metrics}
\label{app:Evaluation Metrics}
we employ Mean Reciprocal Rank (MRR),  Mean Rank, Hits@1, and Hits@10 as evaluation metrics:
\begin{itemize}
    \item \textbf{MRR.} The MRR is a metric used in information retrieval and other fields to evaluate the quality of a ranked list of results. For a set of queries, it calculates the average of the reciprocal of the rank of the first correct answer.
    \item \textbf{Mean Rank.} Mean Rank simply calculates the average rank of the relevant items in the ranked list. It sums up the ranks of all relevant items for a set of queries and divides by the number of queries.
    \item \textbf{Hits@1.} Hits@1 is a metric that measures the proportion of queries for which the correct answer is retrieved as the top-1 candidate. 
    \item \textbf{Hits@10.} Hits@10 is similar to Hits@1, but instead of checking if the correct answer is ranked first, it checks if the correct answer is among the top 10 results in the ranked list. It evaluates how well the system can retrieve the relevant item within the first 10 positions.
\end{itemize}

\begin{table*}[!ht]
    \small
    \centering
    \caption{Exploration experiments of initial number of samples $\alpha$ and growth factor $\beta$ based on CLASS (AMAN) for the text-molecule retrieval task. The blue shaded row indicates the trade-off experimental result determined by max-min regularization. The best results are shown in bold and the second-best results are underlined.}
    \begin{tabular}{c|c|cccc}
    \toprule
        \textbf{$\alpha$} & \textbf{$\beta$} & \textbf{Hits@1($\uparrow$)} & \textbf{Hits@10($\uparrow$)}
        & \textbf{MRR($\uparrow$)} & \textbf{Mean Rank($\downarrow$)} \\ 
        \midrule
        50 & 1 & 49.8  & 91.3  & 0.644  & 20.78 \\
        20 & 2 & 48.7  & 91.3  & 0.636  & 25.26 \\
        40 & 2 & 50.6  & 91.4  & 0.653  & \underline{14.09} \\
        60 & 2 & \textbf{51.4}  & 91.5  & 0.658  & 23.72 \\
        80 & 2 & 50.1  & 92.3  & 0.651  & 22.07 \\
        10 & 3 & 49.7  & 91.5  & 0.645  & \textbf{13.85} \\
        \rowcolor{cyan!10} 40 & 3 & \underline{51.1}  & \textbf{92.6}  & \textbf{0.660}  & 16.80 \\
        70 & 3 & 50.1  & 91.9  & 0.651  & 17.79 \\
        20 & 4 & 49.6  & 92.0  & 0.649  & 15.37 \\
        60 & 4 & 49.7  & 92.1  & 0.646  & 20.80 \\
    \bottomrule
    \end{tabular}
    \label{tab:Parameter exploration text2mol}
\end{table*}

\begin{table*}[!ht]
    \small
    \centering
    \caption{Exploration experiments of initial number of samples $\alpha$ and growth factor $\beta$ based on CLASS (AMAN) for the molecule-text retrieval task. The blue shaded row indicates the trade-off experimental result determined by max-min regularization. The best results are shown in bold and the second-best results are underlined.}
    \begin{tabular}{c|c|cccc}
    \toprule
        \textbf{$\alpha$} & \textbf{$\beta$} & \textbf{Hits@1($\uparrow$)} & \textbf{Hits@10($\uparrow$)}
        & \textbf{MRR($\uparrow$)} & \textbf{Mean Rank($\downarrow$)} \\ 
        \midrule
        50 & 1 & 45.7  & 90.8  & 0.615  & 20.27 \\
        20 & 2 & 45.6  & 90.6  & 0.613  & 24.10 \\
        40 & 2 & 46.8  & 91.3  & 0.625  & \underline{15.00} \\
        60 & 2 & 47.6  & 91.7  & \textbf{0.634}  & 23.36 \\
        80 & 2 & 47.2  & \underline{91.9}  & 0.629  & 21.33 \\
        10 & 3 & 45.7  & 91.2  & 0.617  & \textbf{14.33} \\
        40 & 3 & 46.8  & \textbf{92.0}  & 0.626  & 17.24 \\
        70 & 3 & 46.7  & 91.4  & 0.626  & 17.10 \\
         \rowcolor{cyan!10} 20 & 4 & \underline{47.8}  & 91.6  & \underline{0.631}  & 15.14 \\
        60 & 4 & \textbf{48.1}  & 91.3  & 0.634  & 20.83 \\
    \bottomrule
    \end{tabular}
    \label{tab:Parameter exploration mol2text}
\end{table*}

\section{Comparison Methods}
\label{app:Comparison Methods}
We compare with the following methods to verify the superiority of our methods:
\begin{itemize}
\item \textbf{MLP-Ensemble \cite{edwards-etal-2021-text2mol}, GCN-Ensemble \cite{edwards-etal-2021-text2mol}, All-Ensemble \cite{edwards-etal-2021-text2mol}.} The ensemble
baselines employ MLP-based or GCN-based models to encode molecules for retrieval learning. Each model is initialized with different parameters.
\item \textbf{MLP+Atten \cite{edwards-etal-2021-text2mol}, MLP+FPG \cite{mlp_fpg}.} The MLP+Atten baseline adopts cross-modal attention to capture information and rerank the retrieval results. Consistent with MLP+Atten baseline, MLP+FPG reranks the top results of the MLP model with the FPGrowth algorithm.
\item \textbf{Atomas-base \cite{atmos}.} The model is pretrained on a large-scale dataset where the molecules are represented by SMILES, aligning the two modalities at three granularities via clustering approach. Atomas is trained on the ChEBI-20 dataset using the same configuration as the study.
\item \textbf{Memory Bank \cite{10821722}.} The method propose a memory bank based feature projector to extract modality shared features and calculate four similarities to narrow the distance between these similarity distributions to enhance cross-modal alignment.
\item \textbf{AMAN \cite{10063974}.} The method encode text and molecule with SciBERT and GTN, respectively. Meanwhile, They introduces triplet loss and adversarial loss to refine text and molecule representations, aiming to align cross-modal representation.
\item \textbf{ORMA \cite{10822800}.} ORMA adopts optimal transport facilitates multi-grained alignments between texts and molecules, simultaneously introducing contrastive learning to refine cross-modal alignments at three distinct scales: token-atom, multitoken-motif, and sentence-molecule.
\end{itemize}

\section{Parameter Exploration}
\label{app:Parameter Exploration}
As described in Eq. \ref{ratio}, we introduce $\alpha$ and $\beta$ hyperparameters for determining the number of samples to be introduced at each curriculum stage. Thus, we conduct experiments based on different $\alpha$ and $\beta$ with AMAN as the backbone to investigate the sensitivity of our model. Due to the varying degrees in the performance of the metrics, we unify the results across the indicators via max-min normalization to precisely determine the optimal performance of the model for different tasks. The results are presented in Table \ref{tab:Parameter exploration text2mol} and Table \ref{tab:Parameter exploration mol2text} for the text-molecule retrieval task and molecule-text retrieval task. In text-molecule retrieval task, the results show that the model achieves the best performance when $\alpha$ and $\beta$ are set to 40 and 3 respectively. Similarly, the model reaches its peak performance with $\alpha$ and $\beta$ set to 20 and 4 in the molecule-text retrieval task. Despite not being the best performance across metrics, it achieves an overall trade-off between retrieval accuracy and the arrangement of retrieval results.

\end{document}